\documentclass[11pt,a4paper]{article}
\usepackage[hyperref]{acl2021}
\usepackage{times}
\usepackage{latexsym}

\usepackage{microtype}

\usepackage{tikz}
\usetikzlibrary{bayesnet}
\usepackage{subfloat}
\usepackage{subfig}
\usepackage[linguistics]{forest}
\usepackage{amsmath}
\usepackage{booktabs}
\usepackage{multirow}
\usepackage{amssymb}
\usepackage{xspace}

\makeatletter
\patchcmd{\pgfutilsolvetwotwoleqfloat}
  { \noexpand\pgfmathfloatdivide@}
  {\noexpand\pgfmathfloatdivide@}
  {}{}
\makeatother

\aclfinalcopy 
\def\aclpaperid{1727} 

\newcommand{\bllipmd}{\textsc{BLLIP-md}\xspace}
\newcommand{\blliplg}{\textsc{BLLIP-lg}\xspace}

\title{Structural Guidance for Transformer Language Models}

\author{Peng Qian$^1$ \quad Tahira Naseem$^2$ \quad Roger Levy$^1$ \quad Ramón Fernandez Astudillo$^2$\\
  $^1$ Department of Brain and Cognitive Sciences, MIT \quad
  $^2$ IBM Research\\
  \texttt{ pqian@mit.edu  \quad tnaseem@us.ibm.com} \\ \texttt{rplevy@mit.edu \quad ramon.astudillo@ibm.com}}

\date{}

\begin{document}
\maketitle

\begin{abstract}
Transformer-based language models pre-trained on large amounts of text data have proven remarkably successful in learning generic transferable linguistic representations. Here we study whether structural guidance leads to more human-like systematic linguistic generalization in Transformer language models without resorting to pre-training on very large amounts of data. We explore two general ideas. The ``Generative Parsing'' idea jointly models the incremental parse and word sequence as part of the same sequence modeling task. The ``Structural Scaffold'' idea guides the language model's representation via additional structure loss that separately predicts the incremental constituency parse. We train the proposed models along with a vanilla Transformer language model baseline on a 14 million-token and a 46 million-token subset of the BLLIP dataset, and evaluate models' syntactic generalization performances on SG Test Suites and sized BLiMP. Experiment results across two benchmarks suggest converging evidence that generative structural supervisions can induce more robust and humanlike linguistic generalization in Transformer language models without the need for data intensive pre-training.
\end{abstract}

\begin{figure*}
    \centering
\begin{tikzpicture}
\small
\draw (-2.2, 1.2) node[rectangle, draw=none, anchor=north](){
\begin{forest}
[S [NP [The] [birds] ] [VP [sang] [ADVP ] ] ]
\end{forest}
};

\node (0,0) [anchor=west] {\textsc{$\langle$bos$\rangle$\hspace{0.2cm}} \textsc{nt(s)}\hspace{0.2cm} \textsc{nt(np)}\hspace{0.2cm} The\hspace{0.2cm}  birds\hspace{0.2cm} \textsc{reduce}\hspace{0.2cm} \textsc{nt(vp)}\hspace{0.2cm}  sang\hspace{0.2cm} \textsc{nt(advp)}\hspace{0.2cm} $\cdots$ };
    \draw[|-|] (0.15,-0.4) --  node[below] {$w_0$} (0.85,-0.4);
    \draw[|-|] (3.45,-0.4) --  node[below] {$w_1$} (3.95,-0.4);
    \draw[|-|] (4.3,-0.4) --  node[below] {$w_2$} (4.85,-0.4);
    \draw[|-|] (7.8,-0.4) --  node[below] {$w_3$} (8.35,-0.4);
    \draw[|-|] (1.2,0.4) --  node[above] {$y_{0:1}$} (3.2,0.4);
    \draw[|-|] (4.0,0.4) --  node[above] {$y_{1:2}$} (4.3,0.4);
    \draw[|-|] (5.1,0.4) -- node[above] {$y_{2:3}$} (7.5,0.4);
    \draw[|-|] (8.6,0.4) -- node[above] {$y_{3:4}$} (10,0.4);
    
\end{tikzpicture}
\vspace{-0.6cm}

\subfloat[Vanilla language model]{
\begin{tikzpicture}
\hspace{0.38cm}
\node (rect) at (1,0.5) [draw,rounded corners,thick,minimum width=2.5cm,minimum height=1.4cm] {};
\node at (0,0) [circle,fill,inner sep=1.5pt](i0){};
\node at (0,1) [circle,fill,inner sep=1.5pt](o0){};
\node at (1,0) [circle,fill,inner sep=1.5pt](i1){};
\node at (1,1) [circle,fill,inner sep=1.5pt](o1){};
\node at (2,0) [circle,fill,inner sep=1.5pt](i2){};
\node at (2,1) [circle,fill,inner sep=1.5pt](o2){};

\node at (0,1.8) [] (w0o){\small $w_1$};
\node at (1,1.8) [] (w1o){\small $w_2$};
\node at (2,1.8) [] (w2o){\small $w_3$};

\node at (0,-0.6) [] (w0i){\small $w_0$};
\node at (1,-0.6) [] (w1i){\small $w_1$};
\node at (2,-0.6) [] (w2i){\small $w_2$};

\path (i0) edge[-] (o0);
\path (i0) edge[-] (o1);
\path (i0) edge[-] (o2);
\path (i1) edge[-] (o1);
\path (i1) edge[-] (o2);
\path (i2) edge[-] (o2);
\path (o0) edge[-angle 60] (w0o);
\path (o1) edge[-angle 60] (w1o);
\path (o2) edge[-angle 60] (w2o);
\path (i0) edge[-] (w0i);
\path (i1) edge[-] (w1i);
\path (i2) edge[-] (w2i);
\end{tikzpicture}
\hspace{0.38cm}
}
\hspace{0.12cm}
\subfloat[Parsing as Language Modelling]{
\begin{tikzpicture}
\node (rect) at (2,0.5) [draw,rounded corners,thick,minimum width=4.5cm,minimum height=1.4cm] {};
\node at (0,0) [circle,fill,inner sep=1.5pt](i0){};
\node at (0,1) [circle,fill,inner sep=1.5pt](o0){};
\node at (1,0) [circle,fill,inner sep=1.5pt](i1){};
\node at (1,1) [circle,fill,inner sep=1.5pt](o1){};
\node at (2,0) [circle,fill,inner sep=1.5pt](i2){};
\node at (2,1) [circle,fill,inner sep=1.5pt](o2){};
\node at (3,0) [circle,fill,inner sep=1.5pt](i3){};
\node at (3,1) [circle,fill,inner sep=1.5pt](o3){};
\node at (4,0) [circle,fill,inner sep=1.5pt](i4){};
\node at (4,1) [circle,fill,inner sep=1.5pt](o4){};

\node at (0,1.8) [] (w0o){\small \textsc{nt(s)}};
\node at (1,1.8) [] (w1o){\small \textsc{nt(np)}};
\node at (2,1.8) [] (w2o){\small The};
\node at (3,1.8) [] (w3o){\small birds};
\node at (4,1.8) [] (w4o){\small \textsc{reduce}};

\node at (0,-0.6) [] (w0i){\small \textsc{$\langle$bos$\rangle$}};
\node at (1,-0.6) [] (w1i){\small \textsc{nt(s)}};
\node at (2,-0.6) [] (w2i){\small \textsc{nt(np)}};
\node at (3,-0.6) [] (w3i){\small The};
\node at (4,-0.6) [] (w4i){\small birds};

\path (i0) edge[-] (o0);
\path (i0) edge[-] (o1);
\path (i0) edge[-] (o2);
\path (i0) edge[-] (o3);
\path (i0) edge[-] (o4);
\path (i1) edge[-] (o1);
\path (i1) edge[-] (o2);
\path (i1) edge[-] (o3);
\path (i1) edge[-] (o4);
\path (i2) edge[-] (o2);
\path (i2) edge[-] (o3);
\path (i2) edge[-] (o4);
\path (i3) edge[-] (o3);
\path (i3) edge[-] (o4);
\path (i4) edge[-] (o4);
\path (o0) edge[-angle 60] (w0o);
\path (o1) edge[-angle 60] (w1o);
\path (o2) edge[-angle 60] (w2o);
\path (o3) edge[-angle 60] (w3o);
\path (o4) edge[-angle 60] (w4o);
\path (i0) edge[-] (w0i);
\path (i1) edge[-] (w1i);
\path (i2) edge[-] (w2i);
\path (i3) edge[-] (w3i);
\path (i4) edge[-] (w4i);
\end{tikzpicture}
}
\hspace{0.12cm}
\subfloat[Language models with Structural Scaffold]{
\begin{tikzpicture}
\node (rect) at (1,0.5) [draw,rounded corners,thick,minimum width=2.5cm,minimum height=1.4cm] {};
\node at (0,0) [circle,fill,inner sep=1.5pt](i0){};
\node at (0,1) [circle,fill,inner sep=1.5pt](o0){};
\node at (1,0) [circle,fill,inner sep=1.5pt](i1){};
\node at (1,1) [circle,fill,inner sep=1.5pt](o1){};
\node at (2,0) [circle,fill,inner sep=1.5pt](i2){};
\node at (2,1) [circle,fill,inner sep=1.5pt](o2){};

\node at (0,1.8) [] (w0o){\small $w_1$};
\node at (1,1.8) [] (w1o){\small $w_2$};
\node at (2,1.8) [] (w2o){\small $w_3$};

\node at (0,-0.6) [] (w0i){\small $w_0$};
\node at (1,-0.6) [] (w1i){\small $w_1$};
\node at (2,-0.6) [] (w2i){\small $w_2$};

\node at (0.5,2.4) [] (y0o){\small $y_{0:1}$};
\node at (1.5,2.4) [] (y1o){\small $y_{1:2}$};
\node at (2.5,2.4) [] (y2o){\small $y_{2:3}$};

\path (i0) edge[-] (o0);
\path (i0) edge[-] (o1);
\path (i0) edge[-] (o2);
\path (i1) edge[-] (o1);
\path (i1) edge[-] (o2);
\path (i2) edge[-] (o2);
\path (o0) edge[-angle 60] (w0o);
\path (o1) edge[-angle 60] (w1o);
\path (o2) edge[-angle 60] (w2o);
\path (i0) edge[-] (w0i);
\path (i1) edge[-] (w1i);
\path (i2) edge[-] (w2i);
\path (o0) edge[-angle 60,bend right=25] (y0o);
\path (o1) edge[-angle 60,bend right=25] (y1o);
\path (o2) edge[-angle 60,bend right=25] (y2o);
\end{tikzpicture}
\hspace{-0.1cm}
\begin{tikzpicture}
\node (rect) at (1,0.5) [draw,rounded corners,thick,minimum width=2.5cm,minimum height=1.4cm] {};
\node at (0,0) [circle,fill,inner sep=1.5pt](i0){};
\node at (0,1) [circle,fill,inner sep=1.5pt](o0){};
\node at (1,0) [circle,fill,inner sep=1.5pt](i1){};
\node at (1,1) [circle,fill,inner sep=1.5pt](o1){};
\node at (2,0) [circle,fill,inner sep=1.5pt](i2){};
\node at (2,1) [circle,fill,inner sep=1.5pt](o2){};

\node at (0,1.8) [] (w0o){\small $w_1$};
\node at (1,1.8) [] (w1o){\small $w_2$};
\node at (2,1.8) [] (w2o){\small $w_3$};

\node at (0,-0.6) [] (w0i){\small $w_0$};
\node at (1,-0.6) [] (w1i){\small $w_1$};
\node at (2,-0.6) [] (w2i){\small $w_2$};

\node at (0.5,2.4) [] (y0o){\small $y_{0:1}$};
\node at (1.5,2.4) [] (y1o){\small $y_{1:2}$};
\node at (-0.5,2.4) [] (pad){\small \textsc{$\langle$pad$\rangle$}};

\path (i0) edge[-] (o0);
\path (i0) edge[-] (o1);
\path (i0) edge[-] (o2);
\path (i1) edge[-] (o1);
\path (i1) edge[-] (o2);
\path (i2) edge[-] (o2);
\path (o0) edge[-angle 60] (w0o);
\path (o1) edge[-angle 60] (w1o);
\path (o2) edge[-angle 60] (w2o);
\path (i0) edge[-] (w0i);
\path (i1) edge[-] (w1i);
\path (i2) edge[-] (w2i);
\path (o0) edge[-angle 60,bend left=25] (pad);
\path (o1) edge[-angle 60,bend left=25] (y0o);
\path (o2) edge[-angle 60,bend left=25] (y1o);
\end{tikzpicture}
}
    \caption{Top: Illustration of a partial constituency tree and corresponding transitions. Bottom: unidirectional transformer language model (a) without explicit structural supervision, (b) for modelling generative action parsing sequence, and (c) with structural scaffold for predicting the local incremental parsing state.}
    \label{fig:models}
\end{figure*}
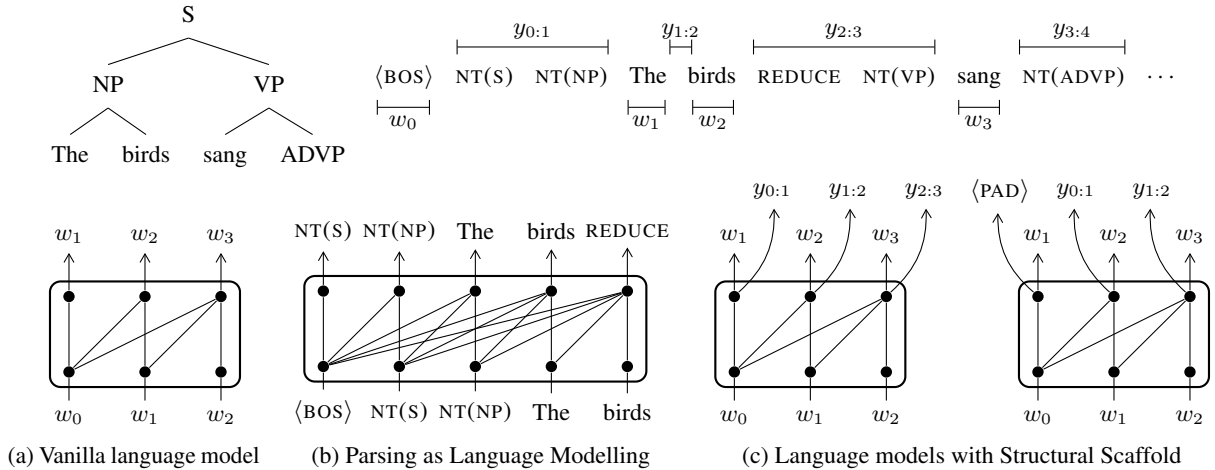

\section{Introduction}

Pre-trained Transformer architectures have led to huge progress in building more human-like language processing systems \citep[among others]{radford2019language,devlin-etal-2019-bert,brown2020language}. These models achieve impressive perplexity results on language modelling datasets, perform well on grammatical judgments \citep{warstadt-etal-2020-blimp-benchmark}, and provide useful linguistic representations that benefit a wide range of downstream tasks. Probing analyses also suggest that these models learn to implicitly encode syntactic information \citep{hewitt-manning-2019-structural,clark-etal-2019-bert} that may support better linguistic generalization than recurrent neural network architectures (RNNs). 

However, the Transformer architecture \cite{NIPS2017_3f5ee243} is an interesting subject of study beyond its success in transfer-learning settings. Transformer models lack the inductive biases of RNNs. Rather than maintaining vector-valued state and updating it in a recurrent manner, auto-regressive Transformer models encode all past decisions simultaneously at each inference step, thanks to a self-attention mechanism. The only notion of sequence order is also given by position embeddings summed to content embeddings in both input and auto-regressive signals.

Previous works have shown the advantage of structural supervision in RNNs in learning to maintain syntactic states and non-local dependencies \citep{kuncoro-etal-2018-lstms,wilcox-etal-2019-structural,futrell-etal-2019-neural}. It remains an open question whether Transformer language models can similarly benefit from generative structural supervision, and what form of structural supervision would more effectively induce human-like syntactic generalization.

This work hypothesizes that the Transformer language model may benefit from explicit generative structural supervision to systematically generalize syntactic knowledge. Here we explore two major classes of structural guidance for Transformer language models based on joint modeling of language and constituency parses. The ``generative parsing as language modeling'' approach builds a Transformer-parameterized model to learn to predict actions that incrementally build constituency trees along with terminal words, following prior work on RNNs~\citep{dyer-etal-2016-recurrent,choe-charniak-2016-parsing}. The ``structural scaffolding'' approach follows the general idea of regularizing hidden representation through multi-task learning objective, with prior success in various NLP tasks \cite{zhang-weiss-2016-stack,sogaard-goldberg-2016-deep,swayamdipta-etal-2018-syntactic}.

We test these two approaches on two subsets of the BLLIP dataset \citep{charniak2000bllip} and evaluate models' syntactic generalization performances on SG Test Suites \citep{hu-etal-2020-systematic} and a sampled subset of the BLiMP Benchmark \citep{warstadt-etal-2020-blimp-benchmark}. We show evidence that generative structural supervision indeed induces more robust and human-like linguistic generalization in Transformer language models and explore the different trade-offs involved in the presented methods.

\section{Models}

Here we explore joint modelling of structures and words parametrized with Transformers by considering both a sentence $W$ and its constituency parse $Y$ and modeling the joint distribution $P(W,Y)$.

\subsection{Generative Parsing as Language Modeling}

A language model can be described formally as a probability distribution over strings of a language $w_1, \cdots, w_T$, usually left-to-right factored.
\begin{equation}
p(W) = p(w_1, \cdots, w_T) = \prod_{t=1}^T p(w_t \mid w_{<t})
\end{equation}
There are many possible approaches that can combine both language modeling and syntax modeling tasks. As long as both tasks share some of the parameters they can be considered a case of multi-task learning \citep{caruana1997multitask}. Of interest here is the model proposed in Recurrent Neural Network Grammars (RNNGs; \citealp{dyer-etal-2016-recurrent}) and parsing as language model (LSTM-LM; \citealp{choe-charniak-2016-parsing}). Both approaches model the joint distribution of words $W$ and constituency tree components $Y$ as 
\begin{equation}
p(Y, W) = p(a_1, \cdots, a_R) = \prod_{t=1}^R p(a_t \mid a_{<t})
\label{eq:palm}
\end{equation}
where $a_t$ are transitions of a state machine that generates both the sentence and the tree. These transitions are similar to the well-established transition sets used for transition-based parsing \citep{earley1970efficient} but adapted to generate both text and parse simultaneously. For the reminder of this work, we will consider each $a_t$ to be integer valued and indexing a dictionary of transitions. A transition $a$ can be a word $w$ or a transition action that generates a component of the constituency tree $y$. The actions include non-terminal symbols that open and label a new constituent with the label $x$, indicated as \textrm{NT(x)}, or a \textrm{REDUCE} action closing the closest open constituent. An example of a partial parse tree and transitions can be found at the top of Figure~\ref{fig:models}.

RNNG and LSTM-LM parametrize the same factorization in Equation~\ref{eq:palm} in different ways. RNNG utilizes stack-LSTMs, which allow it to dynamically create representations for partial tree components by composition. The LSTM-LM, however, uses a flat parametrization treating the transitions as a sequence in a conventional language model learnt with an LSTM \citep{hochreiter1997long}. It should also be noted that the LSTM-LM is designed as a parser, while RNNG is also used as a language model. In order to derive a language model from a joint model, it is is necessary to marginalize over all possible parse trees 
\begin{equation}
p(W) = \sum_{Y \in \mathcal{Y}(W)} p(Y, W) 
\label{eq:marg}
\end{equation}
which is an intractable problem since there is an exponentially large number of possible trees. The original RNNG work \cite{dyer-etal-2016-recurrent} proposes an approximate solution based on importance sampling. In this work we use the word-synchronous beam search approximation introduced in \citet{stern-etal-2017-effective}.

The marginalized likelihood language model in Equation~\ref{eq:marg} is desirable because it makes no statistical independence assumption between language and syntax and shares all parameters across both tasks, with the exception of action specific embeddings. Particularly relevant for this work is the fact that both word and non-word transitions are predicted as language model output indiscriminately and are available at each prediction step through its history $a_{<t}$.

In this work we propose to parametrize Eq~\ref{eq:palm} with a Transformer language model \cite{NIPS2017_3f5ee243}. This is equivalent to the flat parametrization of the LSTM-LM but using a Transformer language model instead. Unlike LSTM-LM, which is a parsing model, we derive from it a language model by marginalization as in the RNNG. A Transformer language model can be succinctly described as a neural network of vertically stacked layers where the $m$-th layer is given by 
\begin{equation}
    \begin{aligned}
    h^{m}_{<t} = \mathrm{FF}^m\left(O \cdot \begin{bmatrix}
    \mathrm{A}^m_1(h^{m-1}_{<t})\\
    \mathrm{A}^m_2(h^{m-1}_{<t})\\
    \cdots \\
    \mathrm{A}^m_N(h^{m-1}_{<t})\\
    \end{bmatrix}\right).
    \end{aligned}
    \label{eq:mha}
\end{equation}
Here $h^{m-1}_{<t} \in \mathbb{R}^{H \times t}$ is the output of the previous decoder layer for all previous predictions of the model at time step $t$ and $H$ is the size of the hidden vector. The input to the first layer i.e. $h^0_{<t}$ are the embeddings of all previous transitions $a_{<t}$ concatenated with a start symbol. Each embedding is the sum of both a content embedding, dictionary vector that is being indexed, and a position embedding that encodes the absolute or relative position of each action in the sequence.

$\mathrm{FF}^m()$ is a feed-forward layer, $A^m_1() \cdots A^M_N()$ are multiple self-attention heads and $O \in \mathbb{R}^{H \times H}$ is a matrix multiplication performed on the concatenated output of the attention heads. Both the feed-forward and the projection of $N$ attention heads through $O$ are wrapped around with residual, dropout and layer normalization operations that are here removed for clarity.  

Each attention head comprises a simple inner product attention mechanism 
\begin{eqnarray}
    \mathrm{A}^m_n(h^{m-1}_{<t}) = V_n^m \cdot h^{m-1}_{<t} \cdot&\nonumber\\  \mathrm{softmax}\big((K_n^m \cdot h^{m-1}_{<t})^T \!\!\!\!\! & \:\cdot Q_n^m \cdot h^{m-1}_{<t} +\mathcal{M}\big)
    \label{eq:ah2}
\end{eqnarray}
where $V_n^m, K_n^m, Q_n^m \in \mathbb{R}^{H/N \times H}$ are value, key and query projection matrices respectively and the \textrm{softmax} operation is normalized over columns to sum to one. The matrix $\mathcal{M} \in \{-\infty, 0\}^{t \times t}$ is used to prevent the model from attending to future states during training, enabling efficient parallelization. It is displayed here due to its relevance for the next section.

Similarly to other models, to derive a distribution over all possible transitions, including words, non-terminal symbols and the \textrm{REDUCE} operation, we can use a softmax together with an inner product
\begin{eqnarray}
	p(a_t \mid a_{<t}) = \mathrm{softmax}(E^{W \cup Y} \cdot h^{m}_{<t})_{a_t}
\end{eqnarray}
where $E^{W \cup Y}$ are the embeddings for the joint vocabulary of words, non-terminals and \textrm{REDUCE} transitions. Henceforth, we refer to this model as \textbf{Parsing as Language Model}, or \textbf{PLM} for short.

Unlike LSTMs or the RNNG, the Transformer has direct access to all past decisions through self-attention and relies on position embeddings to encode word order. Thus, in principle, there is no structural bias for the model to favor past decisions that are close in time to inform current prediction. On one hand, this potential ability to use long distance information can enable a less local, more human like processing of language, but on the other hand, it can also result in an additional learning burden, especially if there is not sufficient learning data available. Also worth noting for the experiments proposed here is that the total number of parameters of a typical Transformer greatly exceeds that of an LSTM or a RNNG model.

\subsection{Incorporating RNNG-like characteristics}

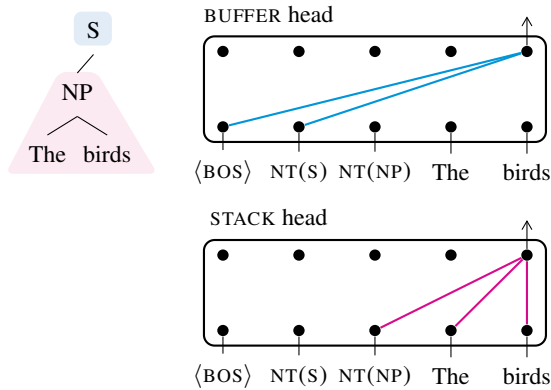
\begin{figure}
    \centering
\hspace{-1cm}
\begin{tikzpicture}
\node at (-4.8,0) {\begin{tikzpicture}
\small
\fill[magenta!20, rounded corners=1mm,opacity=0.5] plot coordinates{(1,-0.45) (0.5,-0.45)
(-0.2, -1.8) (1.8, -1.8)}--cycle;
\fill[cyan!60!blue!20, rounded corners=1mm,opacity=0.5] plot coordinates{(1.2,0.35) (0.7,0.35) (0.7, -0.1) (1.2, -0.1)}--cycle;
\begin{forest}
[S [NP [The] [birds] ] [,phantom] ]
\end{forest}
\end{tikzpicture}};
\node at (0, 0) {\begin{tikzpicture}
\node (rect) at (2,0.5) [draw,rounded corners,thick,minimum width=4.5cm,minimum height=1.4cm] {};
\node at (0,0) [circle,fill,inner sep=1.5pt](i0){};
\node at (0,1) [circle,fill,inner sep=1.5pt](o0){};
\node at (1,0) [circle,fill,inner sep=1.5pt](i1){};
\node at (1,1) [circle,fill,inner sep=1.5pt](o1){};
\node at (2,0) [circle,fill,inner sep=1.5pt](i2){};
\node at (2,1) [circle,fill,inner sep=1.5pt](o2){};
\node at (3,0) [circle,fill,inner sep=1.5pt](i3){};
\node at (3,1) [circle,fill,inner sep=1.5pt](o3){};
\node at (4,0) [circle,fill,inner sep=1.5pt](i4){};
\node at (4,1) [circle,fill,inner sep=1.5pt](o4){};

\node at (4,1.6) [] (w4o){};

\node at (0,-0.6) [] (w0i){\small \textsc{$\langle$bos$\rangle$}};
\node at (1,-0.6) [] (w1i){\small \textsc{nt(s)}};
\node at (2,-0.6) [] (w2i){\small \textsc{nt(np)}};
\node at (3,-0.6) [] (w3i){\small The};
\node at (4,-0.6) [] (w4i){\small birds};

\path (i0) edge[-,draw=cyan!80!blue,thick] (o4);
\path (i1) edge[-,draw=cyan!80!blue,thick] (o4);
\path (o4) edge[-angle 60] (w4o);
\path (i0) edge[-] (w0i);
\path (i1) edge[-] (w1i);
\path (i2) edge[-] (w2i);
\path (i3) edge[-] (w3i);
\path (i4) edge[-] (w4i);

\node at (0.6, 1.5) [draw=none] (labe){\small \textsc{buffer} head};
\end{tikzpicture}};
\node at (0, -2.7) {\begin{tikzpicture}
\node (rect) at (2,0.5) [draw,rounded corners,thick,minimum width=4.5cm,minimum height=1.4cm] {};
\node at (0,0) [circle,fill,inner sep=1.5pt](i0){};
\node at (0,1) [circle,fill,inner sep=1.5pt](o0){};
\node at (1,0) [circle,fill,inner sep=1.5pt](i1){};
\node at (1,1) [circle,fill,inner sep=1.5pt](o1){};
\node at (2,0) [circle,fill,inner sep=1.5pt](i2){};
\node at (2,1) [circle,fill,inner sep=1.5pt](o2){};
\node at (3,0) [circle,fill,inner sep=1.5pt](i3){};
\node at (3,1) [circle,fill,inner sep=1.5pt](o3){};
\node at (4,0) [circle,fill,inner sep=1.5pt](i4){};
\node at (4,1) [circle,fill,inner sep=1.5pt](o4){};

\node at (4,1.6) [] (w4o){};

\node at (0,-0.6) [] (w0i){\small \textsc{$\langle$bos$\rangle$}};
\node at (1,-0.6) [] (w1i){\small \textsc{nt(s)}};
\node at (2,-0.6) [] (w2i){\small \textsc{nt(np)}};
\node at (3,-0.6) [] (w3i){\small The};
\node at (4,-0.6) [] (w4i){\small birds};

\path (i2) edge[-,draw=magenta!90!blue,thick] (o4);
\path (i3) edge[-,draw=magenta!90!blue,thick] (o4);
\path (i4) edge[-,draw=magenta!90!blue,thick] (o4);

\path (o4) edge[-angle 60] (w4o);
\path (i0) edge[-] (w0i);
\path (i1) edge[-] (w1i);
\path (i2) edge[-] (w2i);
\path (i3) edge[-] (w3i);
\path (i4) edge[-] (w4i);
\node at (0.6, 1.5) [draw=none] (labe){\small \textsc{stack} head};
\end{tikzpicture}};
\end{tikzpicture}

\vspace{-0.2cm}

    \caption{Illustration of how the generated incremental constituency parse is used to inform attention patterns in the structure-guided attention heads.}
    \label{fig:attention-mask}
\end{figure}

As previously mentioned, unlike any of the other models, the RNNG is able to create partial tree representations by composition using stack-LSTMs. This changes the RNNG model structure dynamically as a function of the partial parse, a very desirable property to derive syntax-aware representations. Moreover, the fact that Recurrent Neural Networks such as LSTMs summarize all information about previous time steps on two hidden vectors, creates a bottleneck that forces the model to focus on the local state. This is a situation where a syntax-aware representation can provide additional value by enabling the local state to better encompass past structures. We conjecture that a similarly constrained local state might benefit Transformer models in learning linguistic regularities, especially in a limited training data scenario.

In an attempt to capture a similar effect in the Transformer, we explore here the idea of masking some attention heads to reflect the parser state as in the stack-Transformer \citep{astudillo2020transition}. In the stack-Transformer, two attention heads are specialized to attend only to the contents of buffer and stack respectively for dependency and semantic parsing tasks. Here we choose to specialize two heads as well for each layer in Equation~\ref{eq:mha}, as depicted in Fig.~\ref{fig:attention-mask}. One attention head attends to the contents of the last open constituent whereas another head attends all other past decisions not involving that constituent. The rest of the heads are left free as in the original Transformer architecture. To constrain the attention heads, we only need to alter the mask $\mathcal{M}$ in Equation~\ref{eq:ah2} to depend on head index $n$ and past actions $\mathcal{M}_n(a_{<t})$, which results in a negligible computation overhead.

This hard masking makes the model structure change dynamically depending on the partial parse and it forces some heads to focus on the local syntactic state. Nevertheless, unlike the RNNG, it does not create new representations of partial parses that can be composed in a recurrent manner at each time step, and some attention heads can still operate unrestricted. We hypothesize that structure-aware attention mechanism may still help the model achieve better generalization. The symbolic representation induces a strong inductive bias to how the model should use the structure that it generates on the fly. We henceforth refer to this model \textbf{PLM-mask}.

\subsection{Scaffolding by Learning to Predict Local Parse States}

Given the strong coupling between the tasks, the marginal likelihood Transformer language model of the previous section can be expected to be strongly influenced by the additional syntax prediction task. This comes however at a big cost. First, sequences combine both words and non-terminal and reduce transitions, yielding longer sentences than those of a normal language model $R > T$. Furthermore the approximated marginalization is computationally intensive and also introduces an approximation error.

One well-established regime that allows joint modeling of tasks at a low complexity is that of the syntactic scaffold \cite{zhang-weiss-2016-stack,sogaard-goldberg-2016-deep,swayamdipta-etal-2018-syntactic}. Scaffolding adds an additional structure prediction task at one of the layers of the model as a separate layer and only during training. This is a minimally intrusive change since it just branches some hidden vector of the network and computes an additional loss. It also has no influence on test runtime and avoids expensive steps such as marginalization.

However, applying the idea of syntactic scaffolding to our present scenario poses one difficulty. If we use a standard language model predicting words $w$ and predict the non-word symbols $y$ separately, we face the problem that the two sequences have different lengths. To overcome this in a straightforward way, we predict the $n$-gram of non-word actions $y_{t:t+n(t)}$ corresponding to the partial parse synchronous with step $t$ when we predict word $w_t$. We use a secondary softmax layer for this action $n$-gram prediction.
\begin{eqnarray}
	p(y_{t:t+n} \mid y_{<t}) = \mathrm{softmax}(E^{Y^*} \cdot h^{m}_{<t})_{y_{t:t+n}}
    \label{eq:out2}
\end{eqnarray}
Here $E^{Y^*}$ is the vocabulary of all transition $n$-grams excluding words found in the train corpus plus a blank symbol. Note that since Scaffolding operates only at train time, we do not need to worry about generalization of these $n$-grams to test time.

The models are thus trained to minimize the loss function $-\log p(Y,W)$ where
\begin{eqnarray}
p(Y, W) &= \prod_{t=1}^T p(w_t \mid w_{<t}) \nonumber\\
        &+ \prod_{t=1}^T p(y_{t:t+n(t)} \mid w_{<t})
\end{eqnarray}
The scaffold can be set so that the synchronous non-word action $n$-grams $y_{t:t+n(t)}$ are predicted either before (Figure~\ref{fig:models}c, left) or after (Figure~\ref{fig:models}c, right) producing $w_t$. We considered both variants in our experiments to empirically assess their impact on performance. We refer to this model as \textbf{Transformer Language Model with Syntactic Scaffold}, or \textbf{ScLM} in short, and its two versions \textbf{ScLM-past} and \textbf{ScLM-next}, for past and next $n$-gram prediction.

\section{Experiments}

\subsection{Model Training}

All models, including the baseline vanilla language models (\textbf{LM} in short), the syntactic scaffold models, and the generative parsing models, are based on the same architecture of GPT-2 small \citep{radford2019language} (117M parameters, 12 layers, $H=768$) and use the same BPE tokenizer, but with randomly initialized weights. We believe this would give us a fair comparison to pretrained GPT-2 as well, in order to evaluate whether structural guidance helps improve sample efficiency. We implemented all the proposed models using Huggingface's Transformer package \citep{wolf-etal-2020-transformers}\footnotemark\footnotetext{Code available at \url{https://github.com/IBM/transformers-struct-guidance}}. 

As our goal here is to study whether structural guidance helps models learn robust humanlike generalization of syntactic knowledge, we train our model on the BLLIP dataset \citep{charniak2000bllip}, an English newswire style corpus used in \citet{hu-etal-2020-systematic}. This makes the results here more comparable to the results reported in previous work, especially with RNNGs. We train the proposed models and the baseline vanilla Transformer language models on \bllipmd, a 14 million-token corpus, and \blliplg, a 46 million-token corpus, both of which are auto-parsed using a state-of-the-art constituency parser \citep{Kitaev-2018-SelfAttentive}. We used the parsed sentences to generate oracle parsing action sequence for PLM and PLM-mask. We collected a list of word-synchronous parsing action sequences from the train and development oracle of \blliplg and use it to parametrize the action $n$-gram vocabulary of ScLMs trained on both \bllipmd and \blliplg. There are 3756 action $n$-gram types from the corpora, including one padding token and one blank token.

All models were trained with learning rate $10^{-5}$, AdamW optimizer, and minibatch of size 5. We trained the models with multiple seeds within the capacity of our resources, in order to accommodate potential variance. In total, there are three seeds of LM, four of ScLM-past, four of ScLM-next, three of PLM, and three of PLM-mask for \bllipmd, and the same number of seeds of each model type for \blliplg. Models were trained until convergence, as suggested by the loss of the development set during training.

\subsection{Targeted Syntactic Evaluation}

To assess whether a trained model systematically generalizes its syntactic knowledge, we employ targeted syntactic evaluation paradigm \citep{marvin-linzen-2018-targeted}. Specifically, we measure models' performance on two held-out test datasets, a collection of syntactic generalization test suites from \citet{hu-etal-2020-systematic} and BLiMP Benchmark from \citet{warstadt-etal-2020-blimp-benchmark}. These two datasets cover a wide range of English syntactic phenomena.

Tests from \citet{hu-etal-2020-systematic}, which we refer as \textbf{SG Test Suites}, consist of hand-designed test suites for evaluating fine-grained syntactic generalization in incremental processing of a linguistic input. The general method is to compare models' surprisals $p(\textrm{continuation}|\textrm{prefix})$ of grammatical and ungrammatical continuations given certain sentence prefixes. We report the accuracy averaged across SG test suites. BLiMP Benchmark features minimal pairs of a grammatical sentence $W$ and an ungrammatical counterpart $W^*$. To evaluate a model on these minimal pairs, one simply compares the likelihood of $W$ and $W^*$ assigned by the model. 

As is implied by the evaluation methods, we need to marginalize out the structure variables for PLM or PLM-mask models in order to estimate the surprisal of a continuation, given a sentence prefix or the likelihood of a complete sentence. We follow similar setup as in \citet{futrell-etal-2019-neural,wilcox-etal-2019-structural} applying word-synchronous beam search \citep{stern-etal-2017-effective} to find a list $Y_k$ of $k$ incremental parses given a sentence prefix $w_{<t}$. We then sum the joint probability $p(w_{<t}, y_{<t})$ over the list of incremental parses given by the model to approximate the likelihood of $p(w_{<t})$. We set the parse beam size to 100, word-synchronous beam size $k$ as 10, and fast track size of 5. Since the search process can be computationally intensive, the large number of items in BLiMP benchmark poses a computational challenge. We therefore select the first 10\% out of the 1000 items in each of the 67 tests of BLiMP Benchmark. We report the accuracy over the 100 items and refer to this down-sized BLiMP Benchmark as \textbf{BLiMP-10\%}.

\begin{figure}[t]
    \centering
    \includegraphics[width=\linewidth]{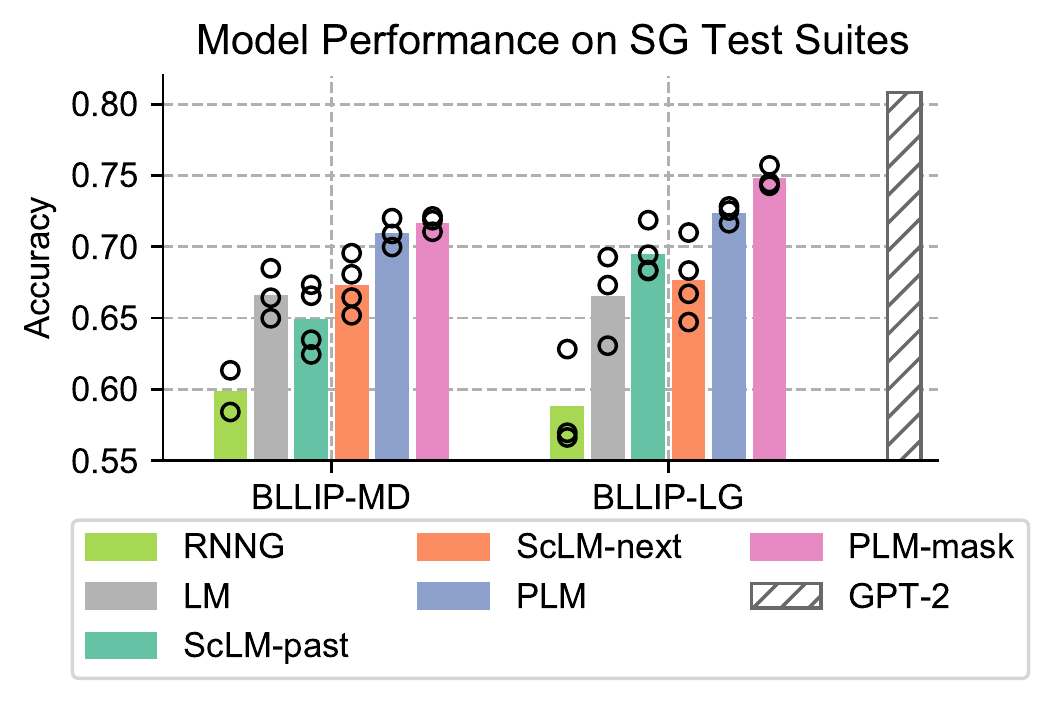}
    
    \includegraphics[width=\linewidth]{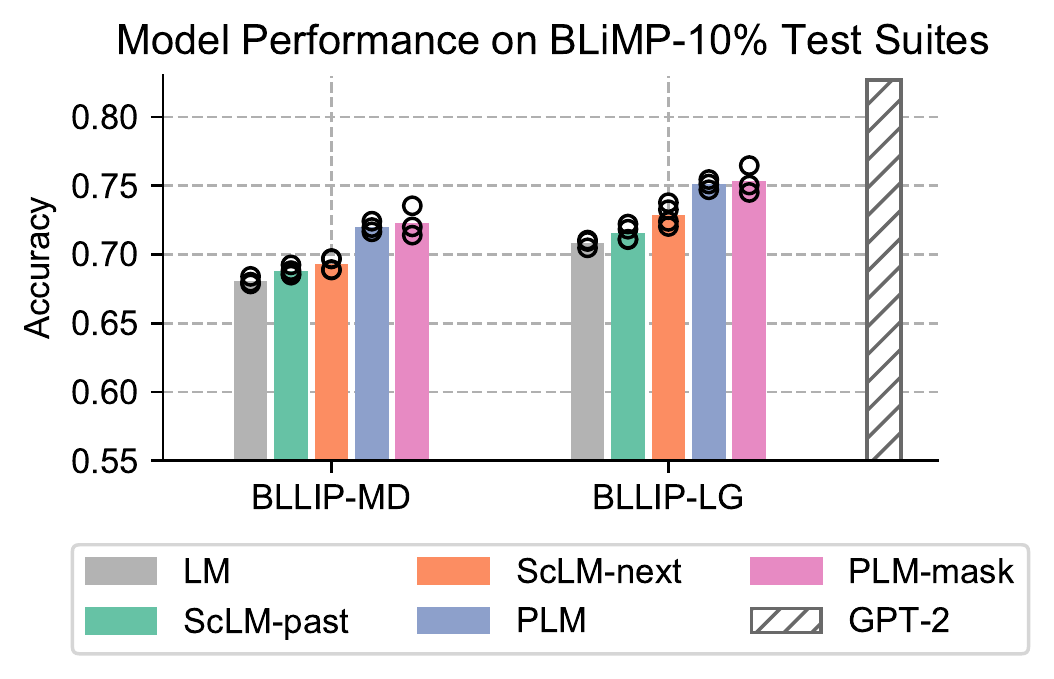}
    \caption{Comparing models' overall accuracy across test suites from SG Test Suites (top) and BLiMP-10\% (bottom). RNNG performances are from \citet{hu-etal-2020-systematic}.
    }
    \label{fig:aggregated-generalization-comparison}
\end{figure}

We compare models' performance on the SG Test Suites and BLiMP-10\% in Figure \ref{fig:aggregated-generalization-comparison}. Each bar shows a model's performance averaged across multiple seeds on a given benchmark, with each dot plotting the accuracy of a specific seed. Overall, syntactic generalization performance improves as the training data size increases from \bllipmd  (14 million tokens) to \blliplg (42 million tokens). Models with structural guidance achieve higher accuracy than the vanilla Transformer language model trained on the same set of raw text data without explicit structural information. We also include the results for the RNNGs taken from \citet{hu-etal-2020-systematic}. RNNG lags behind all Transformer models by a large margin in average scores. We also notice that among different forms of structural guidance, generative parsing as language modeling is the most effective in improving syntactic generalization performance against the baseline transformer language models. We didn't observe consistent benefits of adding dynamic masking mechanism to PLM. While scaffolding approach slightly improves vanilla Transformer language models, it still falls behind the best performance of the model trained with generative parsing. We hypothesize that our scaffold did not fully exploit the compositional structure in the local parses by modelling each action $n$-gram as a distinct type, while the generative parsing models only predict actions in a relatively small set of non-terminal action space, which might make it easier for PLM and PLM-mask to learn compositional generalization. We leave it for future work to design new scaffolds that can take advantage of the combinatorial nature of syntactic structure.

For completeness, we also ran the pre-trained GPT-2 model on the syntactic suites. This yielded a score of 0.808 on the SG Test Suites and 0.827 on BLiMP-10\% for the small version of pre-trained GPT-2. Among models trained on \blliplg, the average accuracy score on the SG Test Suites is 0.723 for PLMs, 0.748 for PLM-masks, and 0.665 for LMs. Similar trend is observed on BLiMP-10\% as well, where among models trained on \blliplg the average accuracy is 0.751 for PLMs, 0.753 for PLM-masks, and 0.708 for LMs.
The proposed PLM method is able to close the gap between GPT-2 small and the same model trained with \blliplg by about half, while the improvement for BLiMP is more modest but still significative. It remains an open question whether scaling syntactic supervision to a larger dataset than \blliplg would bring the generalization performance of PLM models closer to that of the pretrained GPT-2 model.

\begin{figure}[t]
    \centering
    \includegraphics[width=\linewidth]{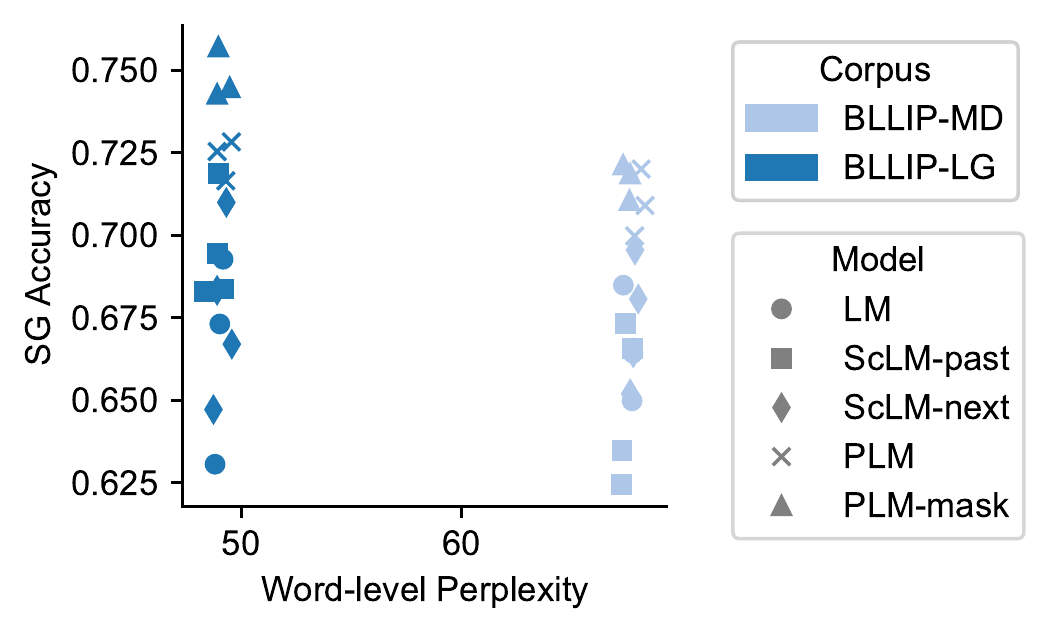}
    
    \includegraphics[width=\linewidth]{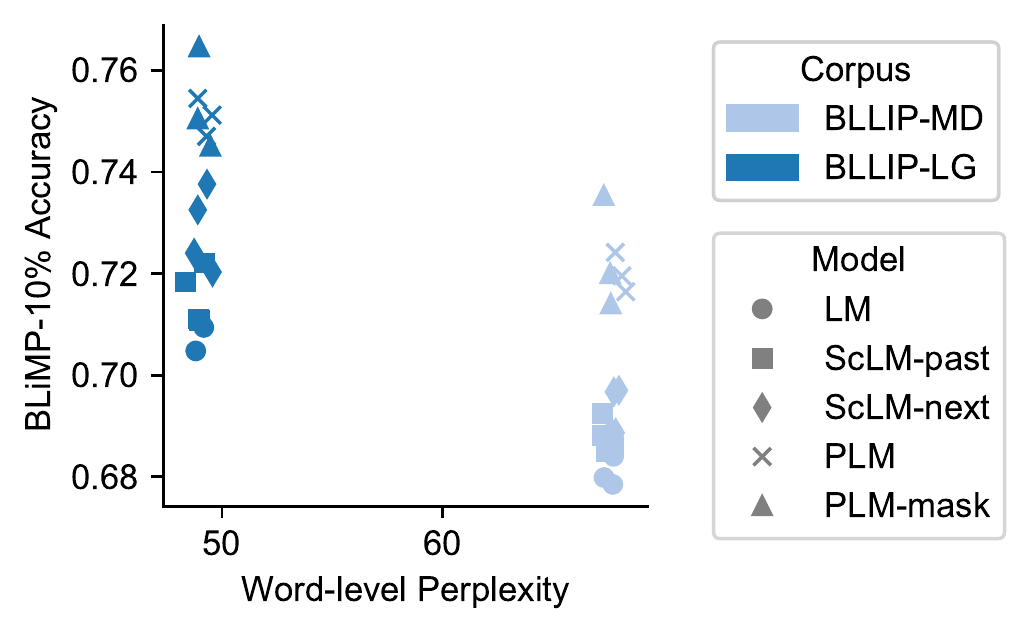}
    \caption{Comparison between model perplexity on BLLIP test data and syntactic generalization performance on SG Test Suites (top) and BLiMP-10\% (bottom).}
    \label{fig:ppl-generalization-comparison}
\end{figure}

\subsubsection{Relationship between Perplexity and Syntactic Generalization Performance}

We compare perplexity on the BLLIP held-out test set against syntactic generalization performance in Figure \ref{fig:ppl-generalization-comparison}. Perplexities of PLM and PLM-mask models are computed setting the parse tree equal to the gold parse in Equation~\ref{eq:marg} to approximate the likelihood. Note that, unlike \citet{hu-etal-2020-systematic}, all our models use the same BPE vocabulary and word tokenization from GPT-2. The only exception are the additional parsing actions in the vocabulary $y$.

From Figure \ref{fig:ppl-generalization-comparison}, both perplexity and syntactic generalization performance improve with dataset size. However, for both training dataset sizes, we see that structural guidance can improve syntactic generalization. PLM models consistently perform better than vanilla models.
While all models achieve very similar perplexity results after being trained on a specific dataset, their syntactic generalization performances differ dramatically.

\begin{figure*}[!t]
    \centering
    \includegraphics[width=\linewidth]{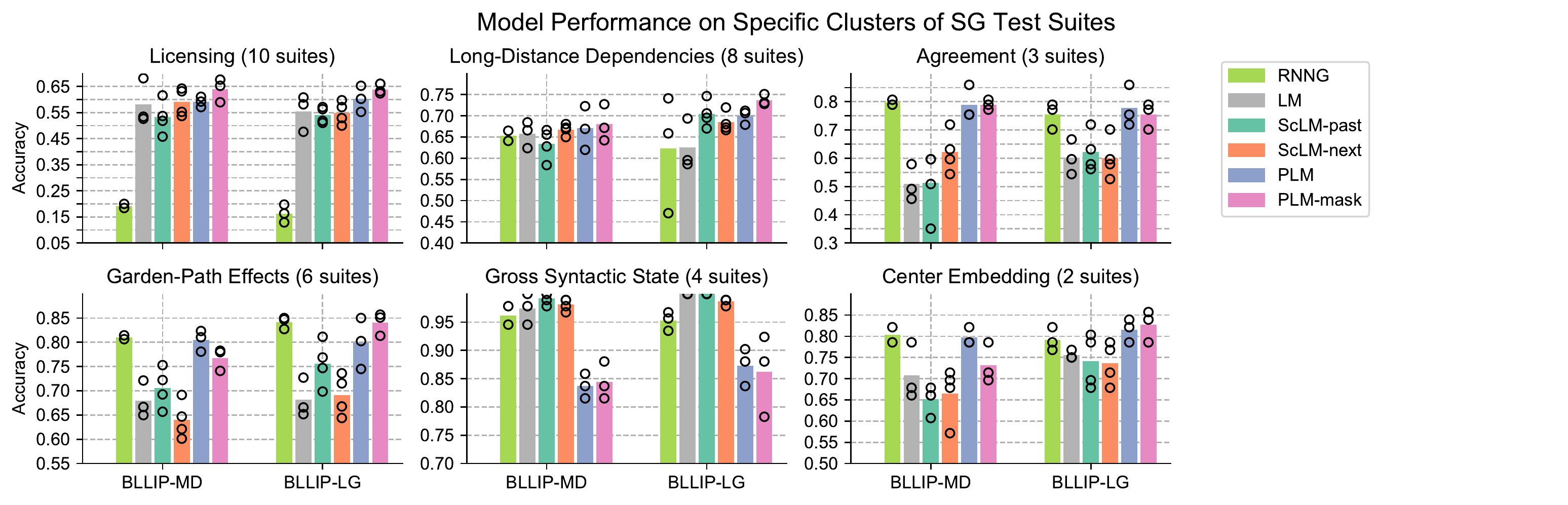}
    
    \vspace{0.25cm}
    \includegraphics[width=\linewidth]{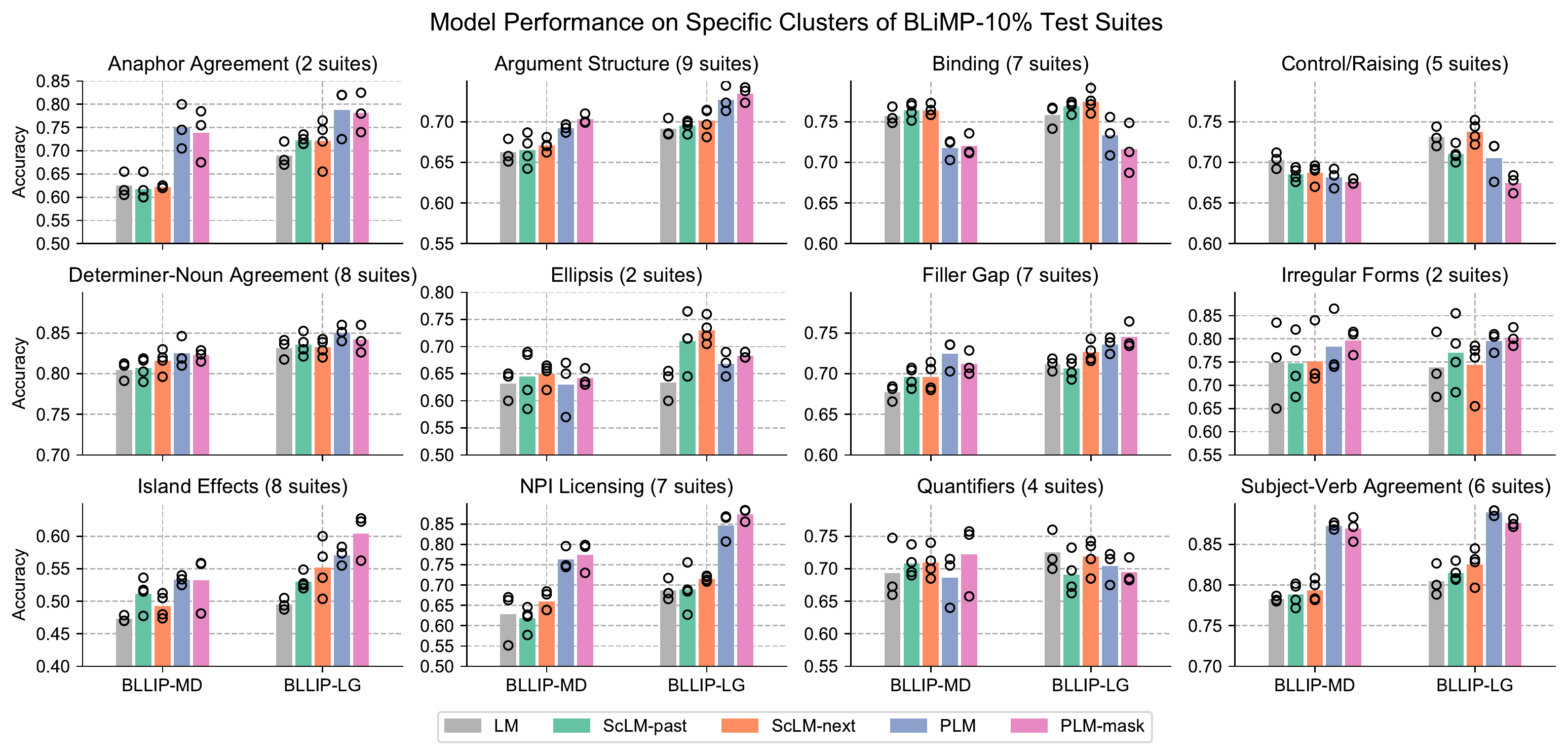}
    \caption{Model performance comparison by specific linguistic phenomena clustered in SG Test Suites (top) and BLiMP-10\% (bottom). RNNG performances are from \citet{hu-etal-2020-systematic}.}
    \label{fig:comparison-by-phenomenon}
\end{figure*}

\subsubsection{Effect of Structural Guidance on Learning Specific Syntactic Structures}

In addition to comparing model's aggregated performances, we also compare their generalization performances in the clustered subsets of tests in SG Test Suites and BLiMP-10\%. These subsets consist of several related tests that target specific type of syntactic phenomenon, such as NPI licensing, subject-verb agreement, filler-gap dependencies, etc. We also include the results for the RNNGs taken from \citet{hu-etal-2020-systematic}. 

Results in Figure \ref{fig:comparison-by-phenomenon} show converging evidence that structural guidance in the form of generative parsing can robustly improve learning of subject-verb agreement and NPI licensing, and helps the model to better capture incremental processing phenomenon such as garden-path effects, but seems to slightly hurt the performance on gross syntactic state. While overall the RNNG shows a poor performance this is mostly due to its very low scores for licensing suites. Excluding these suites only the RNNG shows a performance close to the PLM model, even outperforming it clearly for the gross syntactic state suites. In this category and binding PLM variants seem inferior to all other models.

\section{Related Work}

Multitask learning \cite{caruana1997multitask} has been applied to a variety of NLP tasks with traditional modeling approaches \cite{miller-etal-2000-novel,sutton2005joint,sutton2007dynamic} as well as more recent neural models \cite{collobert2011natural,li2020transformer}. A recurring theme has been the use of structure in the form of syntactic trees to benefit other NLP tasks. Among the early works exploring this direction, \newcite{punyakanok-etal-2008-importance} showed that syntactic parses can benefit Semantic Role Labeling (SRL). \newcite{poon2009unsupervised} extended this idea to induce first-order logic representation in a unsupervised fashion, by clustering the dependency structures. In both cases syntax forms part of a pipeline and is not strictly supervision for the end task. 

This trend continued with the rise of neural models. \newcite{collobert2011natural} improved deep convolution neural network for syntactic chunking models with additional POS supervision. \citet{zhang-weiss-2016-stack,sogaard-goldberg-2016-deep} observe the benefits of POS supervision at  different depths of a neural network model with impact on dependency parsing, tagging and CCG super tagging performance. \newcite{he2019syntax} perform a syntax-based pruning of semantic roles, showing benefits in a multilingual setting. More recently, \newcite{sachan2020syntax} incorporate a syntactic graph recurrent neural network into BERT models for better semantic role labeling. However, their method shows little or no benefit of syntax modeling for Named Entity Recognition and relation linking task. Neural machine translation \cite{chen2018syntax} and text generation \cite{li2020transformer} have also been shown to benefit from syntactic modeling. In a recent work, \newcite{li2020improving} use syntactic modeling in BERT based transformers to achieve performance gains on several text classification benchmarks. Other works have found that structural supervision in the form of intermediate fine-tuning (e.g., on CCG super tagging) is not helpful or even harmful \cite{pruksachatkun2020intermediate,warstadt2019investigating}.

The focus of our work is on gauging the impact of joint modeling on syntactic generalization performance. In this direction, the work of \citet{swayamdipta-etal-2018-syntactic} is close to the scaffolding version of our model. They predict multiple labels, extracted from syntactic information, as auxiliary task and show positive effects on shallow semantic parsing and co-reference resolution. We use however a single feature, constituency parsing $n$-gram, which is closer to prior work relying on Part-of-Speech information. In addition, we explore impact of using preceding structure as feature vs postceding structure, which as shown plays a role in the learning process. 

In terms of modeling objective and syntactic representations, our method is closest to the works of \newcite{choe-charniak-2016-parsing,dyer-etal-2016-recurrent} that jointly model syntax and language. A more recent work from \citet{peng-etal-2019-palm} uses Rational Neural Networks language model that can derive binary unlabeled constituents from attention weights and can supervise the attention to attain a structural inductive bias. The proposed models show lower language modeling perplexity compared to their structure agnostic counterparts. We also extend here the idea of syntax-aware language modeling to transformer-based language models.

Finally, our approach relates to the other works that propose ways of incorporating structural information into Transformer-based models. This includes the use of dependency or tree structure for constraining self-attention patterns \citep{strubell2018linguistically,wang-etal-2019-tree,zhang2020sg}, guiding cross-attention \cite{chen2018syntax,astudillo2020transition}, modelling syntactic distance \citep{du-etal-2020-exploiting}, using syntactic information to guide the computation flow in the model \citep{shen-etal-2021-explicitly}, or through knowledge distillation \citep{kuncoro2020syntactic}. Our structured masking in parsing as language modeling approach is close in spirit to the methods that modify attention mechanism according to syntactic connections \cite{astudillo2020transition}; This work, however, primarily aims to study the impact of structural guidance on syntactic generalization. Therefore, we resort to simpler methods of incorporating structure to minimize the impact of modeling intricacies.

\section{Conclusion}

Our work explores two forms of syntactic supervision as structural guidance for Transformer language models. Experiments suggest that generative parsing approach can effectively improve systematic generalization of learned syntactic knowledge in small training data regime, while a naive syntactic scaffold approach does not improve the baseline to the same extent despite reduced computation cost at inference time. Future work may explore alternative structural guidance strategies that combine the best of both approaches.

\section*{Acknowledgments}

The authors would like to thank the anonymous reviewers for their helpful comments. This work was supported by the MIT-IBM Watson AI Lab.

\bibliographystyle{acl_natbib}
\bibliography{acl2021}

\end{document}